\documentclass[sigconf]{aamas}

\usepackage{balance} 
\usepackage{comment}
\usepackage{subcaption}
\usepackage{algorithm}
\usepackage{algpseudocode}

\setcopyright{none}
    \acmConference[ALA '26]%
    {Proc.\@ of the Adaptive and Learning Agents Workshop (ALA 2026)}%
    {May 25 -- 26, 2026}%
    {Paphos, Cyprus, https://alaworkshop2026.github.io/}%
    {Aydeniz, Delgrange, Mohammedalamen, Yang (eds.)}%
    \copyrightyear{2026}
    \acmYear{2026}
    \acmDOI{}
    \acmPrice{}
    \acmISBN{}
\def\peer/{PEER}
\acmSubmissionID{24}

\title{Preference Guided Iterated Pareto Referent Optimisation\\ for Accessible Route Planning}

\author{Paolo Speziali}
\affiliation{
  \institution{Vrije Universiteit Brussel}
  \city{Brussels}
  \country{Belgium}}
\email{paolo.speziali@vub.be}

\author{Arno De Greef}
\affiliation{
  \institution{Vrije Universiteit Brussel}
  \city{Brussels}
  \country{Belgium}}
\email{arno.de.greef@vub.be}

\author{Mehrdad Asadi}
\affiliation{
  \institution{Vrije Universiteit Brussel}
  \city{Brussels}
  \country{Belgium}}
\email{mehrdad.asadi@vub.be}

\author{Willem R\"opke}
\affiliation{
  \institution{Vrije Universiteit Brussel}
  \city{Brussels}
  \country{Belgium}}
\email{willem.ropke@vub.be}

\author{Ann Nowé}
\affiliation{
  \institution{Vrije Universiteit Brussel}
  \city{Brussels}
  \country{Belgium}}
\email{ann.nowe@vub.be}

\author{Diederik M. Roijers}
\affiliation{
  \institution{Gemeente Amsterdam}
  \city{Amsterdam}
  \country{Netherlands}}
\affiliation{
  \institution{Vrije Universiteit Brussel}
  \city{Brussels}
  \country{Belgium}}
\email{diederik.roijers@vub.be}

\begin{abstract}

We propose the Preference Guided Iterated Pareto Referent Optimisation (PG-IPRO) for urban route planning for people with different accessibility requirements and preferences. With this algorithm the user can interact with the system by giving feedback on a route, i.e., the user can say which objective should be further minimized, or conversely can be relaxed. This leads to intuitive user interaction, that is especially effective during early iterations compared to information-gain-based interaction. Furthermore, due to PG-IPRO's iterative nature, the full set of alternative, possibly optimal policies (the Pareto front), is never computed, leading to higher computational efficiency and shorter waiting times for users.
\end{abstract}


\newcommand{\BibTeX}{\rm B\kern-.05em{\sc i\kern-.025em b}\kern-.08em\TeX}

\begin{document}


\pagestyle{fancy}
\fancyhead{}


\maketitle 


\section{Introduction}
Urban accessibility is increasingly recognized as the key to social participation, economic opportunity, and overall quality of life \cite{pineda2024amsterdamforall}. Yet, for many people with limited mobility, navigating urban environments remains challenging due to limited accessibility. For example, when navigating from say the zoo to the station, one possible route might be too narrow, while another may have too many crossings and has surface issues. Moreover, different people may have different requirements (or just preferences) with regard to such accessibility features \cite{pinhao2022accessible,saha2019project,vanoorschot2024where}

Traditional route planning algorithms, typically formulated as single-objective shortest path problems \cite{dijkstra1959note}, optimize only for distance or travel time and therefore fail to capture the multi-objective nature of accessibility. Therefore, multi-objective shortest path planning problem (MOSPP) formulations \cite{salzman2023heuristic} should be used to model multiple, potentially conflicting (accessibility and other) objectives for routes. Algorithms for MOSPPs typically produce a Pareto front of possible routes, i.e., a set of routes with all possibly optimal trade-offs between the objectives if nothing is known about the preferences of the user. Subsequently, preference elicitation \cite{chen2004survey} methods can then be used to find a suitable route from such a Pareto front \cite{zintgraf2018ordered}. 

However, the requirement to first find the entire Pareto front of optimal solutions to enable preference elicitation is rather sub-optimal. The cost of finding or generating a single Pareto-optimal solution can already become quite high and negatively influence the response time, and consequently the user experience, let alone if the user first has to wait for all possibly optimal routes to be generated. Therefore we pursue an interactive approach that iteratively returns Pareto-optimal solutions faster and, in parallel, solicits lightweight human feedback to guide search toward solutions that increase user satisfaction. 

\begin{figure}[th]
  \centering
  \includegraphics[width=0.65\linewidth]{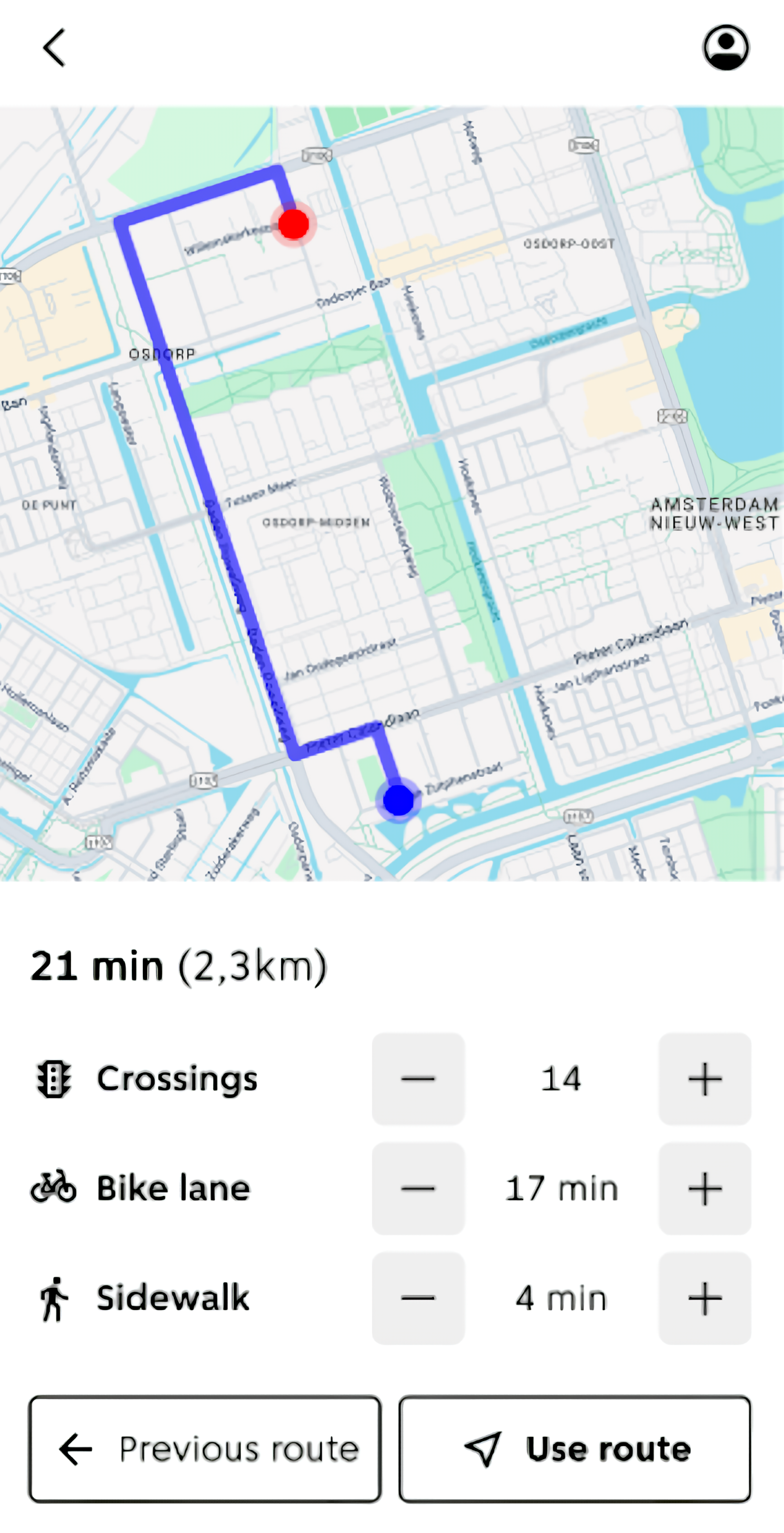}
  \caption{Interface design for interactive route planning.}
  \cite{peer5p2_deliverable}
  \label{fig:design}
\end{figure}
This work is done in the context of the \peer/ project (https://peer-ai.eu)
, and specifically, the Amsterdam accessible route planning use case. In practice this meant that rather than starting from an algorithmic perspective, the municipality of Amsterdam together with the academic partners from the \peer/ consortium \cite{stefanija2024human}
started from user experience design; people with limited mobility have been interviewed about how they would want technology to support them. This has led to a design (Figure~\ref{fig:design}), that was again verified with users. 
The design implies interactions that are rather different than the assumed interactions under some other algorithms, i.e., rather than comparing routes in an order based on information gain (as in~\cite{zintgraf2018ordered}), the user would view a given route (Figure \ref{fig:design}) and then decide which objective should be minimized further (or conversely could be relaxed). A new route is then presented in comparison, the user chooses which route to continue with, and then, if so desired can decide to optimize further. An explicit observation from user interviews that is important in this respect is that users are not willing to interact more than a couple of times with the system to further optimize their route, and will often just pick a satisfactory one. This means that while eliciting preferences, users should not be asked many queries, i.e., the system must improve user satisfaction as quickly as possible after a limited number of interactions. In other words, the primary evaluation criterion for the underlying algorithms is optimization speed in terms of the number of interactions and their response time.

In this paper, we build both on the algorithmic basis of multi-objective optimization, and specifically the IPRO algorithm~\cite{ropke2024divide}, and the City of Amsterdam's design (Figure~\ref{fig:design}) to propose a new algorithm that follows this interaction pattern. We call this algorithm \emph{Preference Guided IPRO (PG-IPRO)}. We compare PG-IPRO both on synthetic data, and on a real accessible routing problem in Amsterdam, to a state-of-the-art preference elicitation method for multi-objective decision making~\cite{zintgraf2018ordered}. We find that PG-IPRO is especially effective to optimize utility in the early iterations of the algorithm, as intended. Furthermore, because due to the iterative nature of PG-IPRO, the algorithm never generates the full Pareto front, resulting in reduced response times.

\section*{Code Availability}

The code base for this paper is available at: \\
\url{https://github.com/plspeziali/pg-ipro}.

\section{Background}

Below we summarize the three background areas most relevant to PG-IPRO: multi-objective shortest path planning problems, to which we apply our work, ordered preference-elicitation strategies used to learn user utilities interactively in multi-objective decision making problems, and the original IPRO algorithm that provides the \textit{divide-and-conquer} framework that we build upon.

\subsection{Multi-objective Shortest Path Planning}
A \textbf{multi-objective shortest path planning problem} (MOSPP) \cite{salzman2023heuristic} is typically defined on a directed graph $G=(V,E)$, where each edge $e=(x_i, x_j) \in E$ between two nodes, $x_i$ and $x_j$, is associated with a vector-valued cost $c(e) \in \mathbb{R}^m$, $m$ being the number of objectives. For a path $P = [x_0, x_1, x_2, ..., x_d]$ connecting a source node $x_0$ to a destination node $x_d$, the overall path value is obtained by aggregating the edge costs along the path, yielding a value vector $v(P) = \sum_{x_i, x_{i+1} \in P} c(x_i, x_{i+1})$. 

Typically, the goal of a MOSPP is not to optimize a single scalar cost but to account for trade-offs between the different objectives represented in $v$. Because objectives may conflict, a single optimal solution may not exist. Instead, solutions are compared using \emph{Pareto dominance}: a path $P_1$ with value vector $v(P_1)$ dominates another path $P_2$ with value vector $v(P_2)$ if $v(P_1)$ is no higher than $v(P_2)$ in all objectives and strictly smaller in at least one objective (assuming minimization). A path is \emph{Pareto-optimal} if no other feasible path dominates it. The set of all Pareto-optimal paths constitutes the \emph{Pareto set}, and their corresponding value vectors form the \emph{Pareto front}, representing all possibly optimal trade-offs among the objectives if all we know of the user is that  less cost is preferred for all objectives but no preference information is available between the objectives.

\subsection{Finding preferred Pareto optimal path}
\label{sec:gppe}

Given a Pareto front, different users will prefer different value vectors and associated paths. To figure out which path a given user prefers we need to have \emph{preference information}. One way of doing this is the interactive multi-objective optimization approaches, often incorporating user feedback to guide the search process \cite{deb2016multi}. In particular, querying users about their preferred direction of improvement allows this optimization process to focus on regions of the trade-off solutions that align with their priorities \cite{thiele2009preference}. In its most basic form, the query reply contains comparisons between two value vectors, i.e., user prefers $v(P_1)$ to $v(P_2)$. However, this is not a very handy format to find the optimally preferred route. One useful approach is to assume that the user preferences can be represented by a (partially observable) utility function $u(v) : \mathbb{R}^m \rightarrow \mathbb{R}$ which maps a value vector to a scalar value. This is called the utility-based approach \cite{hayes_practical_2022}, and implies a total order over value vectors.  

Using the utility-based approach, Zintgraf et al.~\cite{zintgraf2018ordered} propose a framework for interactive preference elicitation aimed at identifying a user’s preferred solution from a set of Pareto-undominated alternatives. Their approach models the decision maker’s latent utility function, $u(\cdot)$, using Gaussian processes (GPs). As such, we refer to this method as \textbf{Gaussian Process Preference Elicitation} (GPPE). It elicits preferences between different value vectors, and learn from this relative feedback. Since human judgments are inherently uncertain and may contain noise due to ambiguity or cognitive limitations, the framework explicitly accounts for noisy preference observations when updating the GP posterior. Rather than relying solely on traditional pairwise comparisons between alternatives, the authors introduce ordered preference elicitation strategies, including ranking- and clustering-based queries that allow users to express richer preference information in each interaction. The system repeatedly proposes candidate solutions with the corresponding vectors and asks the user to compare them; the comparison outcomes are used to update the GP posterior. Note that to select the next vector (e.g., next alternative route), however, the whole Pareto front needs to be generated first, in order to see which next vector would likely yield the most information. Generating such a full Pareto front may be prohibitively expensive in large combinatorial domains, such as route planning on city graphs. In contrast, the new algorithm interleaves a region-focused search with lightweight single-objective optimization to quickly produce user-relevant Pareto solutions, and never computes the entire Pareto front.

\subsection{IPRO}

Rather than generating the complete Pareto front beforehand, we develop our method based on the \textbf{Iterated Pareto Referent Optimisation} (IPRO) algorithm~\cite{ropke2024divide}. IPRO is a divide-and-conquer, anytime algorithm for finding Pareto fronts with a guaranteed accuracy upper bound, by iteratively adding a new solution at every iteration. Below, we discuss the original IPRO, noting that it does not include human feedback.

Given tolerance \(\tau\), IPRO returns an approximation \(V^\tau\) that satisfies
\begin{equation}
\forall \, v \in V^*,\ \exists \, v' \in V^\tau:\ \|v - v'\|_\infty \le \tau,
\end{equation}
where \(V^*\) denotes the set of Pareto-optimal objective vectors.

\begin{figure*}[t!]
    \centering
    \begin{subfigure}[t]{0.33\textwidth}
        \centering
        \includegraphics[width=\linewidth]{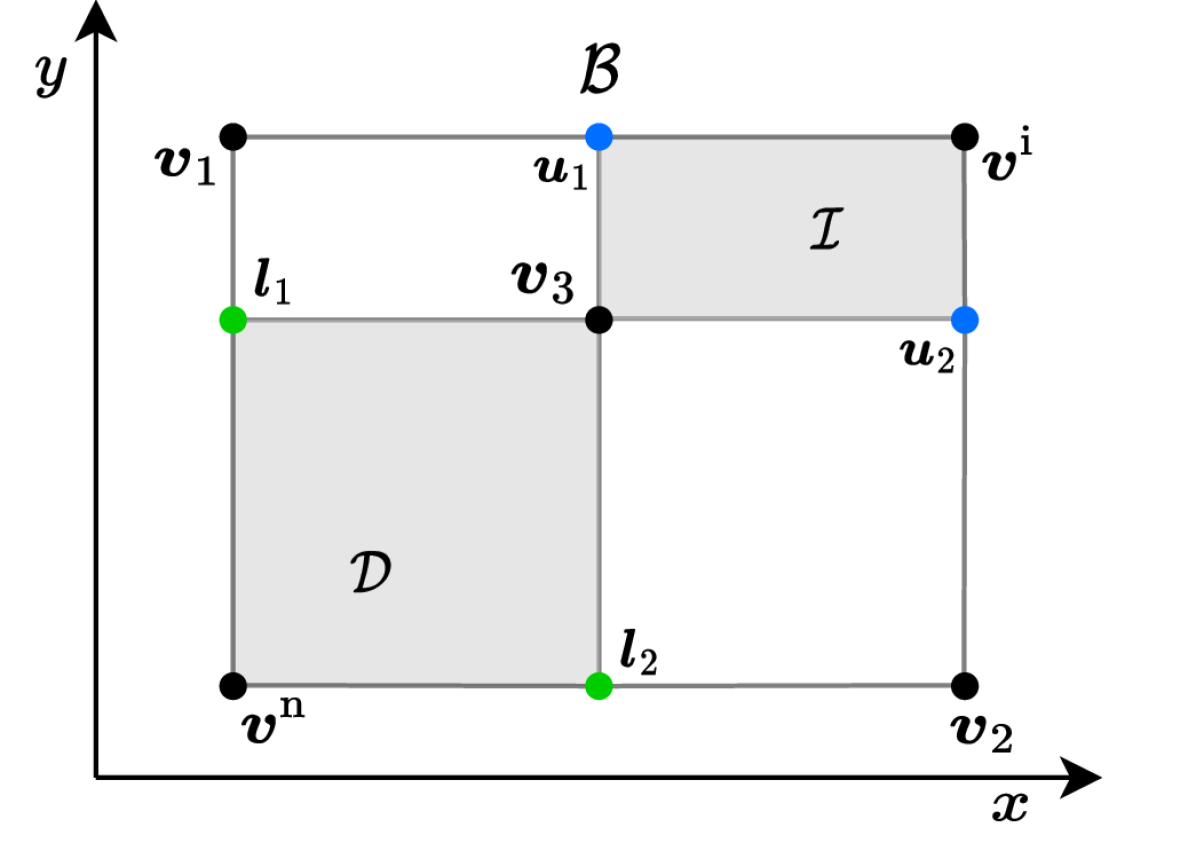}
        \caption{}
        \label{fig:ipro1}
    \end{subfigure}%
    ~ 
    \begin{subfigure}[t]{0.33\textwidth}
        \centering
        \includegraphics[width=\linewidth]{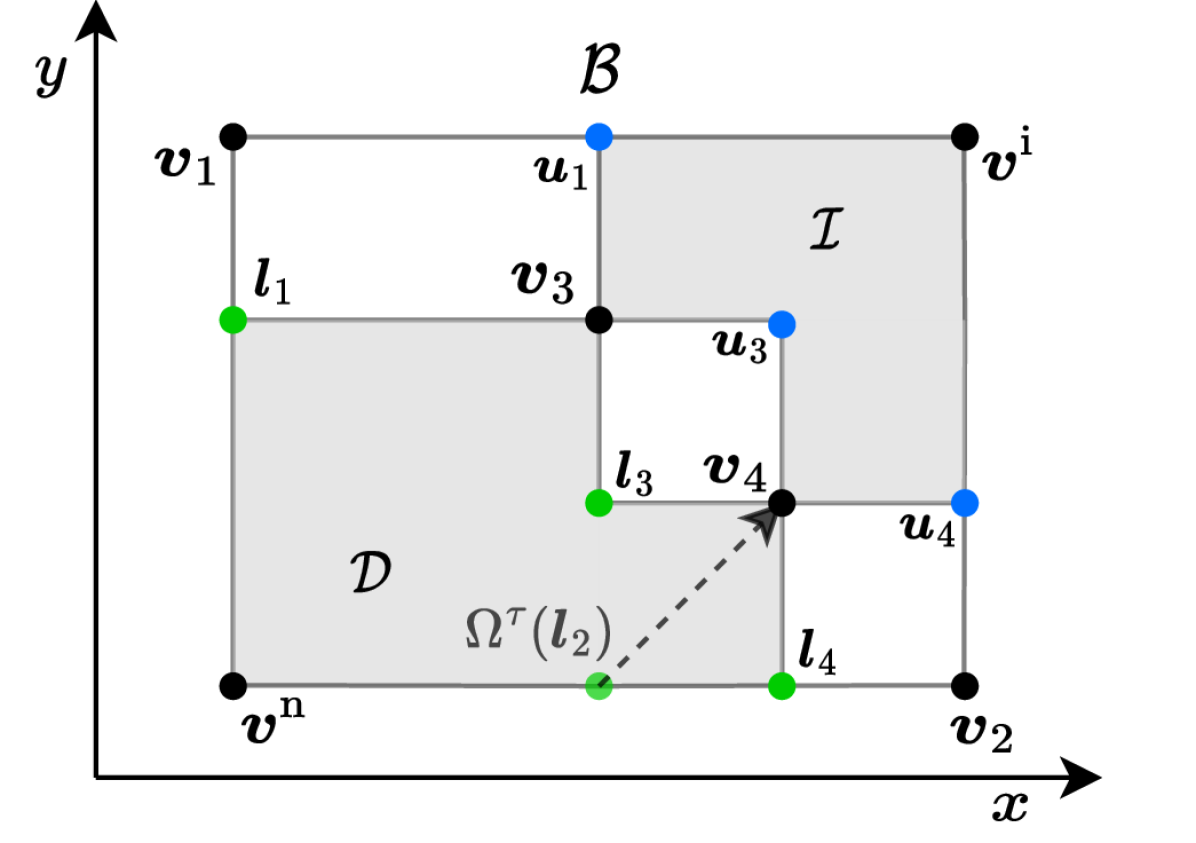}
        \caption{}
        \label{fig:ipro2}
    \end{subfigure}
    ~ 
    \begin{subfigure}[t]{0.33\textwidth}
        \centering
        \includegraphics[width=\linewidth]{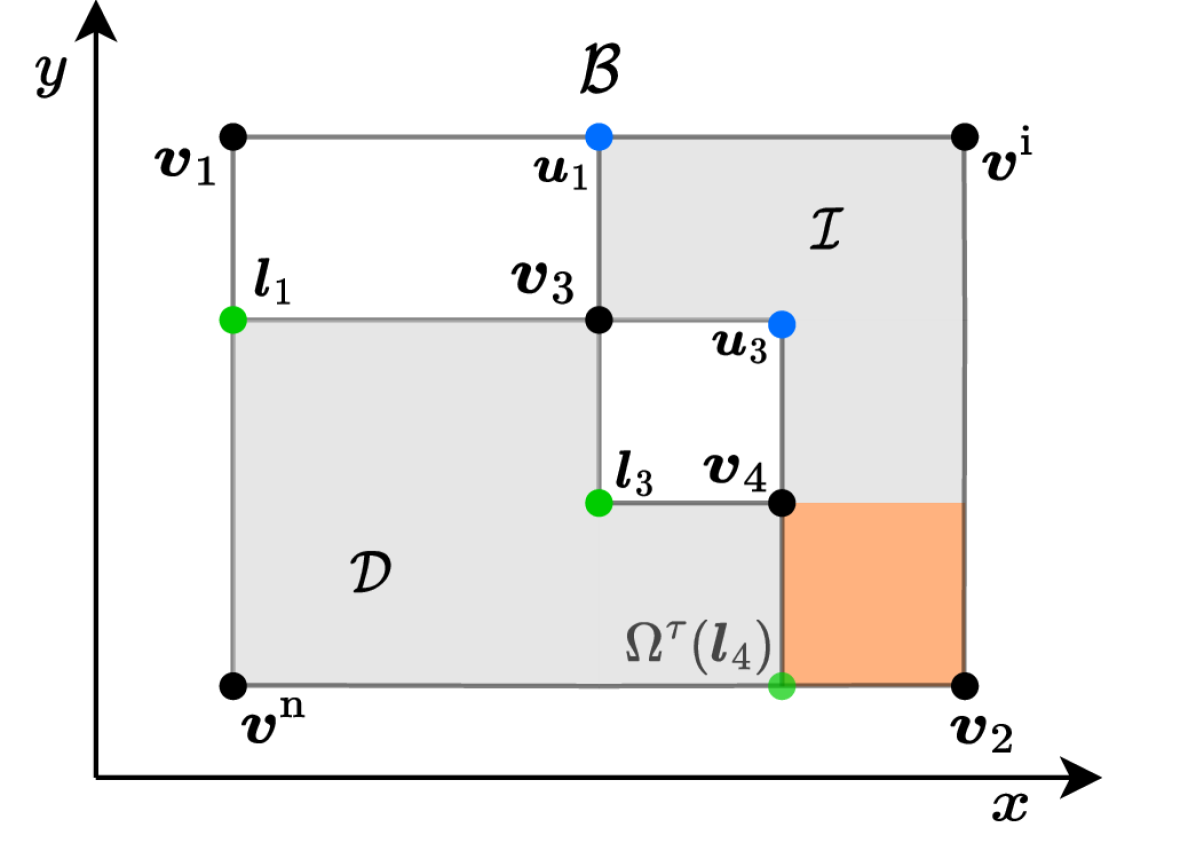}
        \caption{}
        \label{fig:ipro3}
    \end{subfigure}
    \caption{Iteratively update the objective search space in IPRO (Figure from \cite{ropke2024divide})}
\end{figure*}

IPRO decomposes the multi-objective search into a sequence of constrained single-objective subproblems, as can be seen in Figure~\ref{fig:ipro1}. The algorithm maintains an objective search box (bounding box) \(\mathcal{B}\) whose outer bounds are initialized from single-objective optima: the \textbf{Ideal} vector \(v^i\) (the per-objective optima) and an initial \textbf{Nadir} estimate \(v^n\). These extrema form the initial Pareto extrema and define the first partitioning of \(\mathcal{B}\). Computing the exact Nadir point can be challenging, as it may require solving longest path-type subproblems in route planning, which are NP-hard in general. Therefore, practical implementations typically rely on approximations.

Each IPRO iteration selects a lower bound (referent) and calls a Pareto Oracle on the corresponding target region (the sub-box defined by that referent and the current upper bounds). The Oracle either returns a Pareto-optimal vector that strictly Pareto-dominates the referent, or reports that no such vector exists in the target region. The returned solution updates \(\mathcal{B}\) by carving the region into:
\begin{itemize}
  \item a dominated set \(\mathcal{D}\) (solutions now known to be dominated),
  \item an infeasible set \(\mathcal{I}\) (regions that cannot contain Pareto-optimal solutions), and
  \item remaining unexplored target regions (which may still contain Pareto-optimal vectors).
\end{itemize}
In Figure~\ref{fig:ipro2} we can see a new portion being added to both \(\mathcal{D}\) and \(\mathcal{I}\) after adding the point \(v_4\) to the Pareto front.

IPRO continues selecting referents and calling the Oracle until all target regions are resolved up to the tolerance \(\tau\). Note that an Oracle can fail for a region (no strictly-dominating solution exists), in that case IPRO marks the region infeasible and continues with other referents (as shown in Figure~\ref{fig:ipro3}).

The IPRO framework is attractive because it decomposes a complex multi-objective search into multiple constrained single-objective subproblems and because it is algorithm-agnostic: the inner procedure may use either a standard single-objective solver or, alternatively, a specialized Oracle (for example, a multi-objective path planner).

However, IPRO's original formulation does not incorporate explicit human interaction during the search process. The selection of referent points and target regions is entirely algorithm-driven and independent of a decision maker’s evolving preferences. As a result, the algorithm may expend computational effort exploring regions of the objective space that are irrelevant to the user, while offering no lightweight mechanism for steering the search toward more desirable trade-offs.

\section{Method}
\label{sec:pg-ipro}
Building off IPRO, in this section we propose  human-in-the-loop algorithm to quickly steer multi-objective optimization to regions with high user utility. The algorithm will adhere to the interaction scheme co-designed with target users explained in the introduction (Figure \ref{fig:design}).

\subsection{Preference-Guided IPRO (PG-IPRO)}
We propose PG-IPRO, which builds on the iterative scheme of IPRO to allow a human to steer the selection of referents, and thereby quickly home in on regions of the search space likely preferred by the user: after IPRO initialization (ideal and initial Pareto extrema), PG-IPRO proposes an initial Pareto-optimal solution and then enters an interactive loop where in each iteration:

\begin{enumerate}
  \item The user indicates which objective they want to see improved (or conversely can be relaxed), i.e., the \emph{preferred objective};
  \item PG-IPRO selects a referent (lower bound) from IPRO's available referents that guarantees improvement in the chosen objective;
  \item PG-IPRO calls the Oracle restricted to that target region;
  \item The Oracle returns a Pareto-optimal solution that strictly Pareto-dominates the referent (if one exists), PG-IPRO then presents it to the user and updates IPRO's bounding-box partitioning (dominated / infeasible / unexplored regions).
\end{enumerate}

\subsection{Referent selection}

Given all the available referents \(\mathcal{R}\), a current Pareto-optimal solution vector \(s\) and a user-favored objective \(i\) to be further optimized, the \emph{improving set} is defined as:
\begin{equation}
    \mathcal{R}_{\text{improve}} = \left\{ r' \in \mathcal{R} \mid r'_i \leq s_i \right\}.
\end{equation}
Two heuristics are used to select the next referent \(r_{\text{next}}\):
\paragraph{Closest-distance heuristic}
\begin{equation}
    r_{\text{next}} = \arg\min_{ r' \in \mathcal{R}_{\text{improve}}} \left \| r' - s \right \|_1,
\end{equation}
yields small, local improvements in the preferred objective.
\paragraph{Middle-distance heuristic}
Let \(o\) be the ideal Pareto-optimal solution optimized for the objective \(i\). Compute the midpoint \((o+s)/2\) and select
\begin{equation}
    r_{\text{next}} = \arg\min_{ r' \in \mathcal{R}_{\text{improve}}} \left \| r' - \frac{o + s}{2} \right \|_1,
\end{equation}
which biases search away from tiny greedy steps and toward more balanced exploration.

\subsection{Oracle: MO-DFS with pruning and upper-bound guidance} 
\label{sec:mo-dfs}
PG-IPRO uses a Multi-Objective Depth-First Search (MO-DFS) Oracle with Pareto-dominance pruning and guidance by scalarized distance-to-target. The Oracle returns a Pareto-optimal path strictly Pareto-dominating the given referent, or reports failure for the region.
Here are the key components:

\begin{itemize}
    \item \textbf{Lower bound preprocessing}: For each objective, compute a node-wise lower bound, i.e., the shortest-path distance from every node to the target with respect to that objective. In our implementation, these bounds are obtained by running Dijkstra's algorithm on the reversed graph.
    \item \textbf{MO-DFS stack search}: Initialize a stack with the source node $s$, its cumulative cost vector $\mathbf{g}(s)=\mathbf{0}$, and a visited dictionary that stores, for each node, the set of non-dominated cost vectors discovered so far.
    Iteratively pop a state $(u, \mathbf{g}(u))$ from the stack and:
    
    \begin{itemize}
        \item Perform a \textbf{goal check}: if $u$ equals the target node and $\mathbf{g}(u)$ strictly Pareto-dominates the current lower bound (i.e., the best known feasible solution in the target region), update the current best solution.
    
        \item \textbf{Loop over neighbors}: for each neighbor $v$ of $u$:
        \begin{itemize}
            \item Compute the cumulative path cost from the source to $v$ via $u$:
            \[
            \mathbf{g}(v) = \mathbf{g}(u) + \mathbf{c}(u,v),
            \]
            where $\mathbf{c}(u,v)$ is the multi-objective edge cost.
            
            \item Compute a lower-bound estimate $\mathbf{h}(v)$ from $v$ to the target and set
            \[
            \text{estimate} = \mathbf{g}(v) + \mathbf{h}(v).
            \]
    
            \item \textbf{Prune} the neighbor if:
            \begin{itemize}
                \item $\exists j$ such that $\text{estimate}_j > \text{current\_best}_j$, or
                \item $\exists j$ such that $\text{estimate}_j = \\\text{upper\_bound\_of\_target\_region}_j$ \\ (strict domination required), or
                \item $\mathbf{g}(v)$ is Pareto-dominated by a previously stored cost vector for node $v$.
            \end{itemize}
    
            \item If not pruned, compute a scalarized \textbf{distance} of the estimate to the ideal or referent point and append $(v, \mathbf{g}(v))$ to a neighbor list together with its priority value.
        \end{itemize}
    
    \end{itemize}
    
    After processing all neighbors, sort the neighbor list by the distance heuristic (smallest first) and push them onto the stack in that order, preserving depth-first exploration with heuristic prioritization.
\end{itemize}

PG-IPRO is able to use two different scalarization-based distance measures to prioritize neighbor expansion:
\paragraph{Manhattan Distance Function}
\begin{equation}
    \text{distance} = \sum_{i=1}^{n} \frac{|\hat{c}_i - t_i|}{l_i - t_i},
\end{equation}

where \(\hat{c}_i\) is the estimated total cost for objective \(i\), \(t_i\) is the target objective value and \(l_i\) the lower bound.

\paragraph{Chebyshev Achievement Scalarization Function (ASF)}
\begin{equation}
    \text{distance} = \max_{i=1}^{n} \left( \frac{|\hat{c}_i - t_i|}{u_i - t_i} \right),
\end{equation}
where \(u_i\) is the upper bound (or ideal) used for normalization.

The Manhattan metric focuses on closeness to ideal, while ASF emphasizes balance across objectives.

\subsection{Interactive loop}
PG-IPRO maintains the current most-preferred solution and proceeds as shown in Algorithm~\ref{alg:pgipro}.

First, IPRO is initialized by executing the \texttt{init\_phase()}, which computes the ideal and nadir points and constructs the bounding box $\mathcal{B}$. An initial Pareto-optimal solution is proposed to the user, either an extreme solution or the first solution returned by the Oracle using the nadir as referent.

Then, a preference elicitation loop is started. At each iteration, the user selects an objective $i$ and a direction of improvement. The internal IPRO routine selects a referent from the improving region and calls the Oracle to search for a new Pareto-optimal solution. If a new solution is found, the bounding box partition is updated accordingly. If no improving solution exists in the chosen direction, the algorithm reports that no further improvement is possible.

The newly generated solution is then pairwise compared with the currently most-preferred solution. If the user prefers the new solution, it becomes the updated most-preferred solution. The process continues until the user terminates the interaction, and the most-preferred solution is returned.

\begin{algorithm}
\caption{Preference-Guided IPRO (PG-IPRO)}
\begin{algorithmic}[1]

\State Initialize IPRO, compute ideal, nadir and the bounding box 
\State $x^{curr} \gets$ first Pareto-optimal solution
\State $x^{best} \gets x^{curr}$

\While{true}
    \If{user exits} \textbf{break} \EndIf
    \State User selects objective $i$ and direction $d$ ($+$, $-$)
    
    \State $x^{new} \gets$ IPRO-Internal-Loop$(i,d)$

    \If{$x^{new} = \emptyset$}
        \State \textbf{return} $x^{best}$
    \EndIf

    \State Update IPRO partition and bounding box
    \State $x^{curr} \gets x^{new}$

    \State User compares $x^{curr}$ with $x^{best}$
    \If{$x^{curr}$ preferred}
        \State $x^{best} \gets x^{curr}$
    \EndIf

\EndWhile

\end{algorithmic}
\label{alg:pgipro}
\end{algorithm}

\section{Experiments}

We compare PG-IPRO to GPPE (seen in Subsection~\ref{sec:gppe}) as a baseline. The experimental goals are twofold: (i) measure how quickly each method improves user utility under a limited query budget, and (ii) quantify the computational trade-offs between an incremental, interactive search (PG-IPRO) and a front-first Bayesian approach (GPPE).

\subsection{Metrics and reporting}
We report the following metrics, averaged across simulated users (trials):
\begin{itemize}
  \item \textbf{Solution utility:} the (noisy) user utility of the solution presented after each query. Utilities are in $[0,1]$.
  \item \textbf{Maximum utility:} the best utility observed up to (and including) each query.
\end{itemize}

\subsection{General experimental setup}
\begin{itemize}
  \item \textbf{Query protocol}: Each trial represents a single (simulated) user sampled by a unique utility function. Each user answers a bounded number of pairwise comparison queries (GPPE: pairwise only; PG-IPRO: one single-objective "which objective to improve?" query followed by the same pairwise comparison). The optimization method then proposes the next Pareto-optimal candidate.
  \item \textbf{User model}: utility is a weighted sum of monotonically decreasing stacked sigmoids (one per objective); noise is added to utility outputs (default $\sigma_{noise}=0.01$).
  \item \textbf{Budget and sampling}: experiments use different budgets per scenario. The best-found (maximum) utility is recorded after each query.
  \item \textbf{Oracle / search heuristics}: the IPRO oracle uses a Chebyshev scalarization heuristic and a middle-distance referent selection for PG-IPRO unless otherwise noted.
\end{itemize}

\subsection{Experiment A: Comparing Pareto front shape (Convex versus Concave)}
In order to test the effectiveness of PG-IPRO against GPPE on different types of Pareto fronts, we first test the algorithms on synthetically constructed Pareto fronts. For this experiment, we construct two artificial bi-objective Pareto fronts of equally distanced value vectors on a convex resp.\ concave curve. Each Pareto front consists of 30 non-dominated artificial value vectors. Instead of calculating routes in an MOSPP, the algorithms simply retrieve the best value vectors from the Pareto front.\\
\textbf{Setup.}
We use the following parameters for the measurement:
\begin{itemize}
  \item Queries per simulated user: up to 15,
  \item Number of users (trials): 300,
  \item Utility noise: 0.01,
  \item Oracle search: Chebyshev distance,
  \item Referent selection (PG-IPRO): middle-distance.
\end{itemize}
\textbf{Results}
The per-query average solution utility and the per-query average \emph{maximum} utility across trials for both PG-IPRO and GPPE were measured.

\begin{figure}[t]
  \centering
  \includegraphics[width=\linewidth]{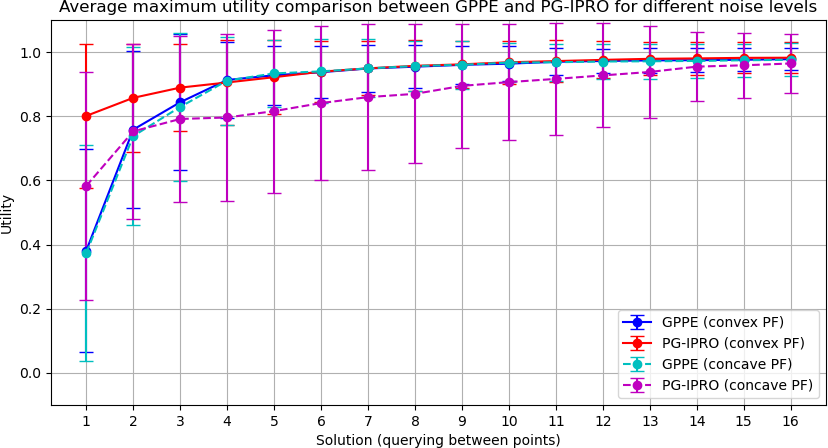}
  \caption{Average maximum utility comparison between PG-IPRO and GPPE for a convex and a concave Pareto front consisting of 30 solutions (noise=0.01)}
  \label{fig:avg_max_util_convexity}
\end{figure}

In Figure~\ref{fig:avg_max_util_convexity} we show the maximum utility comparison between PG-IPRO and GPPE.
In the first queries PG-IPRO achieves a higher average maximum utility than GPPE. This reflects PG-IPRO's anytime, single-solution search: it can present a competitive candidate quickly and then exploit lightweight, directed user guidance to increase satisfaction early in the interaction.
After the initial queries, the information-based GPPE algorithm catches up, and becomes almost equal in utility.

As we noted in the introduction, this is a promising result for PG-IPRO. The good performance after only a few queries is a highly desirable feature for the problem at hand, where high performance in the early queries is key.

\subsection{Experiment B: Comparing Route Planning Bi-Objective Problem Scenarios in Osdorp-Midden}
\begin{figure}[ht]
  \centering
  \includegraphics[width=0.85\linewidth]{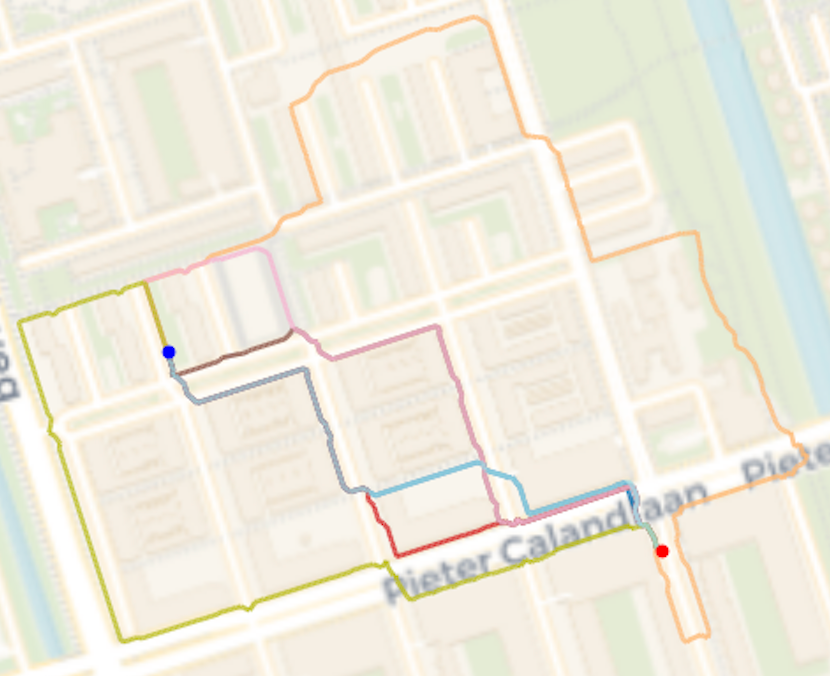}
  \caption{The seven possibly optimal routes in Osdorp-Midden. The blue marker (above) indicates the origin and the red marker (below) the destination. Each colored path corresponds to one Pareto-undominated solution with objective vector $[d,c]$, where $d$ denotes route length (in metres) and $c$ the number of crossings. The yellow-green route on the extreme left corresponds to $[928,3]$, while the orange route extending furthest to the right corresponds to $[1335,2]$. In the central region, the pink and brown routes in the upper part correspond to $[703,4]$ and $[603,5]$, respectively, and the red route in the lower central area corresponds to $[586,6]$. The two routes closest to the lower boundary are the dark blue route $[568,8]$ and the light blue route $[574,7]$.}
  \label{fig:sevenroutes}
\end{figure}

To evaluate the algorithms on realistic accessible route planning problems, we run the algorithms on a problem instance from Amsterdam \cite{gemeente_amsterdam_2024_14198542}. Specifically, we use a given origin and destination point in the Osdorp-Midden district of Amsterdam, and two objectives; the route length (in meters) and the number of crossings. The full Pareto front for this instance has 7 routes as seen in Figure \ref{fig:sevenroutes}. 
\newline
\textbf{Setup.}
For this problem, PG-IPRO uses MO-DFS (Section \ref{sec:mo-dfs}) as the single-route subroutine. The following parameters are used for the measurement:
\begin{itemize}
  \item Pareto front: 7 routes,
  \item Queries per simulated user: up to $6$ (realistic small budget),
  \item Number of users (trials): $50$ (route generation is computationally more expensive),
  \item Utility noise: $0.01$,
  \item Oracle search: Chebyshev distance,
  \item Referent selection (PG-IPRO): middle-distance.
\end{itemize}
\textbf{Results.}
We measure the average utility per query, the average maximum utility per query and the generation time comparison: (1) the time to compute the entire approximated Pareto front (for GPPE) versus (2) the average time to compute a single Pareto-optimal route (for PG-IPRO) were measured.

\begin{figure}[ht]
  \centering
  \includegraphics[width=\linewidth]{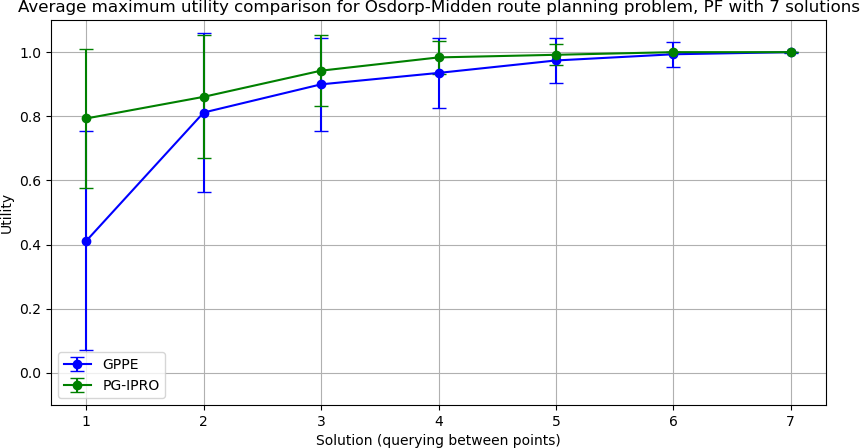}
  \caption{Average maximum utility for Osdorp-Midden PF with 7 solutions (noise=0.01)}
  \label{fig:avg_max_util_OM7}
\end{figure}

In Figure~\ref{fig:avg_max_util_OM7} we present the maximum utility comparison between PG-IPRO and GPPE and we can observe this pattern:
PG-IPRO often attains the highest average maximum utility within the first few queries, so minimal subsequent guidance rapidly locates a highly satisfactory route.
GPPE shows more exploration among enumerated PF points early on; because the PF is small, GPPE can still find high-utility solutions but typically needs a couple more queries to match PG-IPRO's early maxima.
However, a key difference between PG-IPRO and GPPE is the time that the user has to wait for solutions. GPPE requires pre-computation of the entire Pareto front, which took 70$s$. On the other hand, PG-IPRO generates solutions iteratively, which took $4.7s$ on average. 

In conclusion, PG-IPRO achieves better performance, especially in the earlier queries. Furthermore, the user has to wait less to see suitable solutions. We thus conclude that for the use-case of accessible route planning, PG-IPRO appears to be the preferable algorithm. 

\section{Conclusion and Future Work}
Planning a route is fundamentally a multi-objective decision-making challenge. We therefore introduced PG-IPRO, an interactive multi-objective optimization method that integrates human-in-the-loop feedback directly into the search process. By guiding exploration within the objective space of Pareto-optimal solutions, PG-IPRO reduces the need to precompute the entire Pareto front and enables targeted discovery of relevant solutions to a few seconds on average. With the comparison against a competitive baseline, we have shown experimentally that our method achieves better performance to maximize user utility, especially in the earlier queries.

While the current study demonstrates the effectiveness of PG-IPRO under controlled experimental settings using synthetic linear utility functions, which may not fully capture the complexity of real-world preference structures, in the future, we plan to deploy PG-IPRO in real-user studies and collect actual interaction data within the framework of the \peer/ project. Another line of research is the formulation of PG-IPRO with some prior assumptions about the distribution of possible utility functions. Incorporating such prior information could significantly improve PG-IPRO's sample efficiency and convergence speed, leading to scalability in higher-dimensional objective spaces. Furthermore, we believe that PG-IPRO could be used for other agent-based multi-objective decision making problems as well, including aligning AI to human ethical standards \cite{vamplew2018human}, multi-objective coordination \cite{roijers2015computing}, and multi-objective heuristic optimization \cite{deb2016multi}. Finally, we believe that accessible route planning algorithms, like PG-IPRO, can be used in combination with the simulation of populations of people with different accessibility needs, to identify bottlenecks in cities, and make fairer choices while planning and maintaining the traffic network in cities \cite{michailidistransiting}.
\balance

\begin{acks}
This paper is part of the PEER project, which has received funding from the European Union’s Horizon Europe Research and Innovation Programme, under Grant Agreement number 101120406. The paper
reflects only the authors’ view and the EC is not responsible for any
use that may be made of the information it contains. Please see our project page for more information: \url{https://peer-ai.eu/en/}. The basis for this work was the MSc thesis of AG \cite{arnothesis}, carried out at Vrije Universiteit Brussel, after which PS further improved the method/code and wrote most of the paper. WR provided the code-base for the original IPRO algorithm, and helped during the research. MA, AN, and DMR supervised the work, provided feedback, and wrote parts of the paper. We acknowledge the work of our partners at the City of Amsterdam for both the design research on accessible route planning, as well as the planning graphs for realistic accessible route planning in Amsterdam. We further acknowledge Luisa M.\ Zintgraf for making the GPPE code freely available.
\end{acks}



\bibliographystyle{ACM-Reference-Format} 
\bibliography{references}


\end{document}